%% file: main-arxiv.tex
\RequirePackage[root]{preamble/modulatex}
\documentclass[english]{IEEEtran}
\subimport{preamble}{shared}
\usepackage{listings}
\usepackage{natbib}
\usepackage{graphicx}

\RequirePackage{hyperref}
\RequirePackage{xcolor}
\definecolor{customBlue}{HTML}{0064bd}
\definecolor{customRed}{HTML}{9F393D}
\hypersetup{
  colorlinks,
  linkcolor=cyan!60!black,
  citecolor=green!40!black,
  urlcolor=customRed
}

\begin{document}

\title{On the impossibility of non-trivial \\ accuracy under fairness constraints}

\author[1,4]{Carlos Pinzón}
\author[1,4]{Catuscia Palamidessi}
\author[2,3]{Pablo Piantanida}
\author[2,4]{Frank Valencia}
\affil[1]{Inria, France}
\affil[2]{CNRS, France}
\affil[3]{Laboratoire des Signaux et Systèmes (L2S), CentraleSupélec, Université Paris-Saclay}
\affil[4]{LIX, École Polytechnique, Institut Polytechnique de Paris}

\maketitle

\begin{abstract}
One of the main concerns about fairness in machine learning (ML) is that, in order to achieve it, one may have to trade off some accuracy. To overcome this issue, Hardt et al. proposed the notion of equality of opportunity (EO), which is compatible with maximal accuracy when the target label is deterministic with respect to the input features.

In the probabilistic case, however, the issue is more complicated: It has been shown that under differential privacy constraints, there are data sources for which EO can only be achieved at the total detriment of accuracy, in the sense that a classifier that satisfies EO cannot be more accurate than a trivial (i.e., constant) classifier. 
In our paper we strengthen this result by removing the privacy constraint. Namely, we show that for certain data sources, the most accurate classifier that satisfies EO is a trivial classifier. Furthermore, we study the trade-off between accuracy and EO loss (opportunity difference), and provide a sufficient condition on the data source under which EO and non-trivial accuracy are compatible.
\end{abstract}


\subimport{body}{introduction.tex}

\subimport{body}{related-works.tex}

\subimport{body}{preliminaries.tex}

\subimport{body}{polygon-short.tex}

\subimport{body}{impossibility.tex}

\subimport{body}{probabilism.tex}

\subimport{body/characterization}{characterization.tex}

\subimport{body}{conclusion.tex}

\bibliographystyle{plainnat}
\bibliography{bibliography/references}


\clearpage
\renewcommand\appendixname{Supplementary Material}
\appendix

\def\section#1{\subsection{#1}}
\let\theLabel\label
\def\label#1{\theLabel{supp-#1}}
\let\theRef\ref
\def\ref#1{\theRef{supp-#1}}

\section{Remark on conditional expectation}\label{appendix:conditional}
The notation $\EE{Y}{X=x,A=a}$ for defining $q(x,a)$ is not an expectation conditioned on the possibly null event $(X=x,A=a)$.
Instead, it is syntactic sugar for the conditional expectation function.

Formally speaking, the function $q$ is not necessarily unique.
It is uniquely defined at the points $(x,a)$ where $\P{X=x,A=a}>0$.
In the other points, it is defined almost everywhere uniquely in a conditional expectation sense, so that for any other valid conditional expectation function $q'$, we have $q(X,A)=q'(X,A)$ almost surely.
Throughout the paper, we prioritize studying the discrete case to avoid this extreme level of formalism without loosing rigor.

\subimport{body}{polygon.tex}

\subsection{Impossibility theorem}

\subimport{body}{theorem-plane.tex}
\subsection{Characterization of the impossibility of non-trivial accuracy}

\subimport{body}{theorem-nta.tex}

\section{Lemmas}

\subimport{body}{lemmas.tex}

\clearpage
\section{Python code}
Algorithm~\ref{alg:plane} and figure generation for reproducibility:
\begingroup
\tiny
\begin{lstlisting}[language=python]
#!/usr/bin/env python
# coding: utf-8

# # Algorithm 1 and Figures
#
# Requirements:
#
# ```
# pip3 install numpy scipy matplotlib SciencePlots
# ````

import itertools
import numpy as np
import matplotlib.pyplot as plt
from matplotlib.patches import Polygon
from matplotlib.collections import PatchCollection
from scipy.spatial import ConvexHull, convex_hull_plot_2d

plt.style.use('science')  # pip3 install SciencePlots

class RandInterval(np.random.RandomState):
    '''Uniform sampling from (a, b)'''

    def open(self, a, b):
        assert a < b
        c = a
        while c == a:
            c = a + self.random() * (b - a)
        return c

def vectorGenerator(seed=None):
    '''Algorithm 1 of the paper. Asserts the 5 conditions'''
    R = RandInterval(seed)
    Q1 = R.open(0, 1 / 2)
    Q2 = R.open(1 / 2, 1)
    Q3 = R.open(1 - Q1, 1)
    P3 = R.open(1 / 2, 1)
    a = max((1 - P3) * Q1, 1 / 2 - P3 * Q3)
    b = min((1 - P3) * Q2, P3 * Q1)
    om = R.open(a, b)
    P2 = (om - Q1 * (1 - P3)) / (Q2 - Q1)
    P1 = 1 - P3 - P2
    P = np.array([P1, P2, P3])
    Q = np.array([Q1, Q2, Q3])
    try:
        # Verify the 5 conditions explicitly
        assert 0 < P1 < 1 and 0 < P2 < 1 and 0 < P3 < 1
        assert 0 < Q1 < 1 and 0 < Q2 < 1 and 0 < Q3 < 1
        assert P1 + P2 + P3 == 1
        assert P1 * Q1 + P2 * Q2 + P3 * Q3 > 1 / 2
        assert Q1 < 1 / 2 and Q2 > 1 / 2 and Q3 > 1 / 2
        assert Q3 + Q1 > 1
        assert P1 * Q1 + P2 * Q2 - P3 * Q1 < 0
    except AssertionError as e:
        e.args = (f'\nP={P.round(3)}\nQ={Q.round(3)}',)
        raise
    return P, Q

def arbitraryGenerator(seed=None, n=4):
    '''Arbitrary generator for explanation purposes'''
    R = np.random.RandomState(seed)
    P = R.random(n)
    P /= P.sum()
    Q = R.random(n)
    return P, Q

def test_generator(many_times):
    '''Test (run) the vectorGenerator many times'''
    assert many_times > 0
    for _ in range(many_times):
        vectorGenerator(seed=None)
    print(f'{many_times} tests passed')
    return

class DataSource:

    def __init__(self, A, P, Q):
        assert len(A) == len(P) == len(Q)
        self.n = len(A)
        self.A = A
        self.P = P
        self.Q = Q

    def print(self):
        print(f'A={self.A.round(3)}')
        print(f'P={self.P.round(3)}')
        print(f'Q={self.Q.round(3)}')

    def err(self, F):
        P = self.P
        Q = self.Q
        return np.sum(F * (1 - 2 * Q)) + np.sum(P * Q)

    def oppDiff(self, F):
        P = self.P
        Q = self.Q
        A = self.A
        f = lambda A: np.sum(F * Q * A) / np.sum(P * Q * A)
        return f(A) - f(1 - A)

    def iter_soft(self, k=11):
        dim = np.linspace(0, 1, k)
        for f in itertools.product(dim, repeat=self.n):
            F = np.array(f) * self.P
            yield F

    def iter_const(self):
        yield self.P * 0
        yield self.P

    def iter_hard(self):
        for f in itertools.product([0, 1], repeat=self.n):
            F = np.array(f) * self.P
            yield F

    def metrics(self, F_iter):
        l = [(self.err(F), self.oppDiff(F)) for F in F_iter]
        return np.array(l)

    def the_plot(self, styles=(), r=None):
        self.print()

        #soft = cls.metrics(self.iter_soft())
        hard = self.metrics(self.iter_hard())
        const = self.metrics(self.iter_const())
        hull = hard[ConvexHull(hard).vertices]

        with plt.style.context(['grid', 'scatter', *styles]):
            #plt.scatter(soft[:,0], soft[:,1], label='Soft', alpha=0.3)
            poly = Polygon(hull, label='Soft', alpha=0.25)
            plt.gca().add_patch(poly)
            plt.scatter(hard[:, 0], hard[:, 1], marker='.', label='Hard')
            plt.scatter(const[:, 0], const[:, 1], marker='x', label='Constant')
            plt.xlabel('error')
            plt.ylabel('opportunity-difference')
            plt.xlim((0, 1))
            if r is not None and r > 0:
                x0 = const[:, 0].min()
                y0 = const[const[:, 0] == x0, 1][0]
                if y0 != 0:
                    print(y0)
                else:
                    plt.xlim((x0 - r, x0 + r))
                    plt.ylim((y0 - r, y0 + r))
            plt.legend()
            plt.show()
        return

test_generator(many_times=10**4)

# Arbitrary non-example for Section "Strong impossibility result"
A = np.array([0, 1, 0, 1])
P, Q = arbitraryGenerator(seed=3, n=4)
DataSource(A, P, Q).the_plot(styles=['ieee'])

# Generated example for Theorem "Impossibility result"
A = np.array([0, 0, 1])
P, Q = vectorGenerator(seed=0)
DataSource(A, P, Q).the_plot(styles=['ieee'])

# First example we found. Appears as "Example 1"
A = np.array([0, 1, 0, 1])
P = np.array([3, 2, 1, 2]) / 8.
Q = np.array([9, 15, 15, 16]) / 20.
DataSource(A, P, Q).the_plot(styles=['ieee'])

# Other arbitrary (not shown in paper)
A = np.array([0, 1, 0, 1, 0, 1, 0, 1])
P = np.array([1, 1, 1, 1, 1, 1, 1, 1]) / 8.
Q = np.array([1, 1, 1, 1, 1, 1, 3, 3]) / 4
DataSource(A, P, Q).the_plot(styles=['ieee'])
\end{lstlisting}
\endgroup

\end{document}

%% file: preamble/shared.tex
%
%
\usepackage{standalone}
\usepackage[T1]{fontenc}
\usepackage{amsmath}
\usepackage{amsthm}
\usepackage{amssymb}
\usepackage{algorithm}
\usepackage{authblk}
\usepackage[noend]{algpseudocode}
\subimport{}{pkg-amsthm.tex}
\subimport{}{pkg-hyperref.tex}

\ifnonroot
\subimport{../bibliography}{pkg-biblatex.tex}
\else
\def\references{}
\fi

\subimport{}{macros-symbols.tex}
\subimport{}{macros-main.tex}
\usepackage{tikz}
\usepackage{tikz-3dplot}
\subimport{}{theme-nonroot.tex}

\makeatletter 
\def\@font@warning#1{}
\makeatother 

\makeatletter
\patchcmd{\@makecaption}
  {\scshape}{}{}{}
\patchcmd{\@makecaption}
  {\\}{.\ }{}{}
\makeatother


%% file: bibliography/pkg-biblatex.tex
\usepackage[style=alphabetic]{biblatex}
\saverelpath[references.bib]{\bibpath}

\addbibresource{\bibpath}
\def\references{
  \ifimported\else\ifcitations
  \printbibliography
  \fi\fi
}

%% file: preamble/macros-symbols.tex
\global\long\def\calE{\mathcal{E}}%
\global\long\def\calQ{\mathcal{Q}}%
\global\long\def\calX{\mathcal{X}}%
\global\long\def\bbE{\mathbb{E}}%
\global\long\def\bbP{\mathbb{P}}%
\global\long\def\bbR{\mathbb{R}}%
\global\long\def\bbZ{\mathbb{Z}}%

\global\long\def\boldE{\boldsymbol{E}}%
\global\long\def\boldP{\boldsymbol{P}}%
\global\long\def\boldOne{\boldsymbol{1}}%

%% file: body/introduction.tex
\section{Introduction}

During the last decade, the intersection between machine learning
and social discrimination has gained considerable attention from the academia, the industry and the public in general.
A similar trend occurred before between machine learning and privacy, and even the three fields have been studied together recently~\cite{pujol2020fair,cummings2019compatibility,kearns2019ethical,agarwal2020trade}.

Fairness, has proven to be harder to conceptualize than privacy, for which differential privacy has become the defacto definition.
Fairness is subjective and laws vary between countries.
Even in academia, depending on the application, the words fairness and bias have different meanings \cite{crawford2017trouble}.
The current general consensus is that fairness can not be summarized into a unique universal definition;
and for the most popular definitions, several trade-offs, implementation difficulties and impossibility theorems have been found~\cite{kleinberg2016inherent,chouldechova2017fair}.
One such definition of fairness is equal-opportunity~\cite{hardt2016equality}.

To contrast equal-opportunity (EO) with accuracy, we borrow the notion of trivial accuracy from the work by Cumming et al. \cite{cummings2019compatibility}.
A \emph{non-trivial} classifier is one that has higher accuracy than any constant classifier.
Since constant classifiers are independent of the input, trivial accuracy determines a very low minimum performance level that any correctly trained classifier should overcome.
Yet as shown in related works \cite{cummings2019compatibility,agarwal2020trade}, under the simultaneous constraints of differential privacy and equal-opportunity, it is impossible to have non-trivially accurate classifiers.  

In this paper, we complement these existing theorems by showing that even without the assumption of differential-privacy, there are distributions for which equal-opportunity implies trivial accuracy.
This is only possible, however, under a probabilistic data source, i.e. when the correct label for a given input is not necessarily deterministic.

Probability plays two different roles in this paper.
On the one hand, we allow classifiers to be probabilistic.
This is important because sometimes randomness is the only fair way to distribute an indivisible limited resource.
Indeed, equal-opportunity and equal-odds demand probabilistic predictors in some scenarios.
On the other hand, we consider the possibility that the data source may be probabilistic.
This decision is motivated by two different reasons:
\begin{enumerate}
  \item It enables a more realistic and general representation of reality: one in which the information in the input may be insufficient to conclude definitely the yes-no decision, or in which real-life constraints force the decision to be different for identical inputs.
  \item It provides a more general, yet simple, perspective for understanding the trade-off between fairness and accuracy.
  Also, it can take into account that in practice, input datasets are a noisy (thus probabilistic) approximation of reality.
\end{enumerate}

Our contributions are the following.
\begin{enumerate}
  \item We prove that for certain probabilistic distributions, no predictor can achieve EO and non-trivial accuracy simultaneously.

\item We provide a sufficient condition that guarantees compatibility between non-trivial accuracy and EO.

\item We explain how to modify existing results that assume deterministic data sources to the probabilistic case:

\begin{enumerate}
  \item We prove that for certain distributions, the Bayes classifier does not satisfy EO.
  As a consequence, in these cases, EO can only be achieved by trading-off some accuracy.
  \item We give sufficient and necessary conditions for non-trivially accurate predictors to exist.
\end{enumerate}

\item We prove and depict several algebraic and geometric properties about the feasible region in the plane of opportunity-difference versus error.

\end{enumerate}

\references

%% file: body/related-works.tex
\section{Related works}

Our paper is strongly related to the following two works that consider a randomized learning algorithm guaranteeing (exact) equal-opportunity and also satisfying differential privacy:
\cite{cummings2019compatibility} shows that, for certain distributions, these constraints imply trivial accuracy.
\cite{agarwal2020trade} proves the same claim for any arbitrary distribution and for non-exact equal-opportunity, i.e. bounded opportunity-difference.
It also highlights that, although there appears to be an error in the proof of \cite{cummings2019compatibility}, the statement is still correct.
In contrast, in this paper, we prove the existence of particular distributions in which trivial accuracy is implied directly from the (exact) equal-opportunity constraint, without any differential privacy assumption.

Another work about differential-privacy and fairness is \cite{pujol2020fair}.
Unlike our work and that of \cite{agarwal2020trade}, which are limited to equal-opportunity (and equal-odds), they use a different notion of fairness, and this allows them to provide examples in which privacy and fairness are not necessarily in a trade-off.

There are also several works that focus on the feasibility of fairness constraints.
In \cite{kleinberg2016inherent}, it is shown that several different fairness notions can not hold simultaneously, except for exceptional cases.
In \cite{lipton2018does}, it is shown that the two main legal notions of discrimination are in conflict for some scenarios.
In particular when impact parity and treatment parity are imposed, the learned model seems to decide based on irrelevant attributes.
These works differ from our work in that they reveal contradictions that arise when different notions of fairness are imposed together.
In contrast, we fix a single fairness notion, equal-opportunity, and reveal a contradiction that may arise with non-trivial accuracy.  

Lastly, our geometric plots differ from those in the seminal paper of equal-opportunity \cite{hardt2016equality}.
Graphically, their analysis is carried out over ROC curves while we plot directly the two metrics of interest.
In this sense, we provide a complementary geometric perspective for analyzing equal-opportunity and accuracy together. 

\references

%% file: body/preliminaries.tex
\section{Preliminaries}

The notation described in this section is summarized in Table~\ref{table:notation}.

We consider the problem of binary classification with a binary protected feature.
\emph{Protected features}, also called sensible attributes or sensible features, are  input features that represent race, gender, religion, nationality, age, or any other variable that could be used to discriminate against a group of people.
A feature may be considered as a protected feature in some contexts and not in others, depending on whether the classification task should ideally consider that feature or not.  
For our purposes, we assume the simple and fundamental case in which there is a single protected attribute that can only take two values, e.g. man or woman, or, religious or non-religious.

\subsection*{Data source}

We consider an observable underlying statistical model consisting of three random variables over a probability space $\Pspace$:
the \emph{protected feature} $A:\Pomega\to\imA$, the \emph{non-protected feature vector} $X:\Pomega\to\imX$ for some positive integer $d$, and the \emph{target label} $Y:\Pomega\to\imY$. 
We refer to this statistical model as the \emph{data source}.

The distribution of $(X,A)$ is denoted by the measure $\xa$ that computes for each ($(X,A)$-measurable) event $E\subseteq \imXA$, the probability $\xa(E) \eqdef \P{(X,A)\in E}$.
To reduce the verbosity of the discrete case, we denote the probability mass function as \(
  \xa(x,a) \eqdef \xa(\set{(x,a)})
\), i.e. \(
  \xa(x,a) = \P{X=x,A=a}
\).

The expectation of $Y$ conditioned on $(X,A)$ is denoted both as the function $q(x,a) \eqdef \EE{Y}{X=x,A=a}$ (conditional expectation, see the supplementary material for more details) and the random variable $Q\eqdef \EE{Y}{X,A} = q(X,A)$.

The random variable $Q$ plays the role of a soft target label because, since $q(x,a) = \PP{Y=1}{X=x,A=a}$, then $Y$ can be modeled as a Bernoulli random variable with success probability $Q$.

The distribution of $(X,A,Y)$ is completely characterized by the pair $(\xa, q)$, hence we refer to this pair as the distribution of the data source.
And we distinguish two cases:
the data source is \emph{probabilistic} in general, but if $Q\in\imY$ (with probability $1$), then it is said to be \emph{deterministic}.
This distinction is crucial, because several statements hold exclusively in one of the two cases.

\begin{table}

\subimport{.}{table-of-symbols.tex}

  \caption{
    \label{table:notation}
    Notation used in the paper.
  }
\end{table}

\subsection*{Classifiers and predictors}

Analogously to the data source, we model the estimation $\Y$ as a Bernoulli random variable with success probability $\Q=\q(X,A)$ for some ($(X,A)$-measurable) function $\q$.
We refer to $\Y$ as a (hard) \emph{classifier}, and to $\Q$ or $\q$ as a (soft) \emph{predictor}.
Notice that $\Y$ is deterministic when $\Q\in\imY$ (with probability 1), in which case, $\Y=\Q$ (w.p. 1).
Hence all deterministic classifiers are also predictors.

The set of all soft predictors is denoted as $\setQ$. We highlight the following predictors in $\setQ$:
\begin{enumerate}
  \item the two constant classifiers, $\hat 0$ and $\hat 1$, given by $\hat 0 (x,a) \eqdef 0$ and $\hat 1(x,a) \eqdef 1$,
  \item for each $\Q\in\setQ$, the $\half$-threshold classifier given by $\thr{\Q}\eqdef\one{\Q>\half}$,
  \item the data source soft target $Q$, and
  \item the Bayes classifier $\thr{Q}=\one{Q>\half}$.
\end{enumerate}

It is well known\footnote{See for instance Chapter~3 of \cite{fukunaga2013introduction}.} that the Bayes classifier $\Bayes$ has minimal error among all predictors in $\setQ$, regardless of whether the data source is deterministic or not.

\subsection*{Evaluation metrics}

To refer to \emph{equal-opportunity} \cite{hardt2016equality}, we introduce a continuous metric called the \emph{opportunity-difference}.
The opportunity-difference of a predictor $\Q\in\setQ$ is defined as \[
\begin{alignedat}{1}
  \oppDiff{\Q}\eqdef & +\PP{\Y=1}{A=1,Y=1}\\
  & -\PP{\Y=1}{A=0,Y=1}
\end{alignedat}
\]
and a predictor $\Q\in\setQ$ is said to satisfy equal-opportunity whenever $\oppDiff{\Q}=0$.

The \emph{error} and the \emph{accuracy} of a predictor $\Q\in\setQ$ are defined as 
\[
\begin{alignedat}{1}
  \err{\Q} & \eqdef\P{\Y\neq Y}\\
  \acc{\Q} & \eqdef1-\err{\Q}
\end{alignedat}
\]

Additionally, we consider a minimal reference level of accuracy that should be outperformed intuitively by any well-trained predictor.
The \emph{trivial accuracy} \cite{cummings2019compatibility} is defined as $\tau \eqdef \max \left\{ \acc{\Q}: \Q\in\setTriv \right\}$, where $\setTriv$ is the set of (trivial) predictors whose output does not depend on $X$ and $A$ at all, and as a consequence is in independent of $Y$ as well.
In other words, $\setTriv$ consists of all constant soft predictors $\setTriv \eqdef \left\{ ((x,a)\mapsto c) : c \in [0,1]\right\}$.
According to the Neyman-Pearson Lemma, the most accurate trivial predictor is always hard, i.e. must be either $\hat 0$ or $\hat 1$.
Thus $\tau$ is well defined and can be computed  as 
\[
  \tau = \max\left\{ \P{Y=0},\,\P{Y=1}\right\}.
\]

A predictor $\Q\in\setQ$ is said to be \emph{trivially accurate} if $\acc{\Q}\leq\tau$, and \emph{non-trivially accurate}, or \emph{non-trivial} otherwise.
Notice that for a degenerated data source in which the decision $Y$ is independent of $X$ and $A$, all predictors are forcibly trivially accurate.

\references

%% file: body/table-of-symbols.tex
\begin{tabular}{|ll|}
  \hline
  $(X,A,Y)$ & Data source\\
  $X$ & Non-protected feature vector in $\bbR^d$\\
  $A$ & Protected feature in $\set{0,1}$\\
  $Y$ & Target label in $\set{0,1}$\\
  $Q,q$ & Soft target label $Q\eqdef \EE{Y}{X,A}$ \\
  $\xa$ & Distribution of $(X,A)$\\
  $(\xa, q)$ & Distribution of $(X,A,Y)$\\
  \hline
  $\Q, \q$ & Predictor $\Q=\q(X,A)=\EE{\Y}{X,A}$\\
  $\Y$ & Predicted label in $\imY$\\
  $\setQ$ & Set of all predictors\\
  $\acc{\Q}$ & Accuracy of $\Q$: $\P{\Y = Y}$\\
  $\oppDiff{\Q}$ & Opportunity difference of $\Q$:\\
   & \small{$\EE{\Q}{Y=1,A=1}-\EE{\Q}{Y=1,A=0}$}\\
  \hline
\end{tabular}

%% file: body/polygon-short.tex
\section{The error vs opportunity-difference region}

In this section, we remark several properties of the region $\regionM\subseteq [0,1]\times[-1,+1]$ given by \[
  \regionM \eqdef \{(\err{\Q},\oppDiff{\Q}): \Q\in\setQ\}
\]
which represents the feasible combinations of the evaluation metrics (error and opportunity-difference) that can be obtained for a given source distribution $(\xa, q)$.

\begin{figure}
  \center
  \includegraphics[width=0.9\columnwidth]{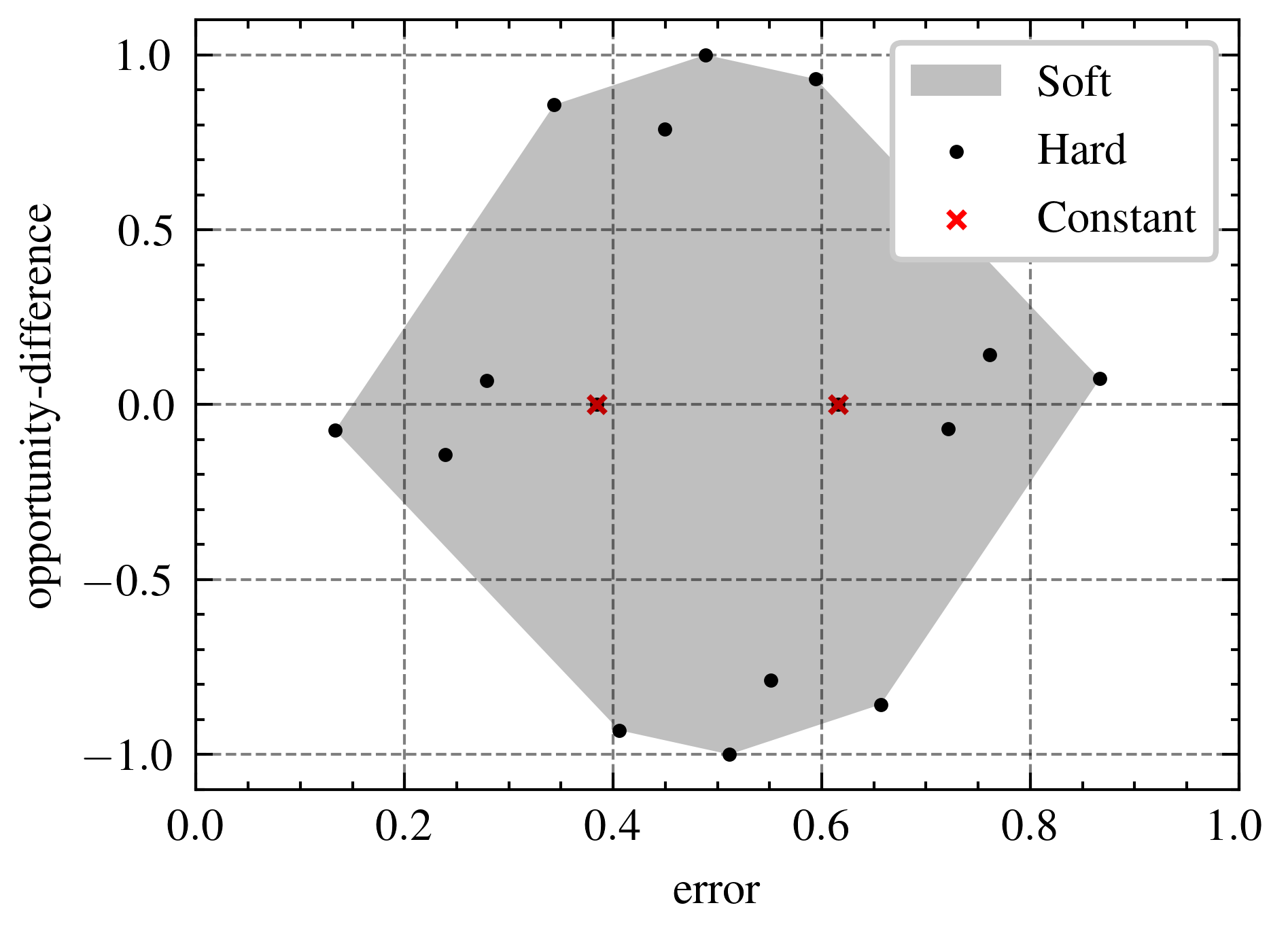}
  \caption{
    \label{fig:non-example}
    Region $\regionM$ for an arbitrary source distribution.
    The data source is not deterministic because the Bayes classifier does not satisfy equal-opportunity.
    The Python code for generating this figure is in the supplementary material.
  }
\end{figure}

$\regionM$ determines the tension between error and opportunity difference.
Figure~\ref{fig:non-example} shows an example of this region.

\begin{theorem}
\label{thm:polygon}
Assuming a discrete data source with finitely many possible outcomes, the region $\regionM$ of feasible combinations of error versus opportunity-difference satisfies the following claims.
\begin{enumerate}
  \item $\regionM$ is a convex polygon.
  \item The vertices of the polygon $\regionM$ correspond to some deterministic predictors.
  \item $\regionM$ is symmetric with respect to the point $(\half, 0)$.
\end{enumerate}
\end{theorem}
\begin{proof}
The proof is in the supplementary material.
It uses the  fact that affine transformations map polytopes into polytopes (See Chapter 3 of \cite{grunbaum2013convex}).
\end{proof}

The reader is invited to visualize the aforementioned properties of $\regionM$ in Figure~\ref{fig:non-example}, which depicts the region $\regionM$ for a particular instance
\footnote{Namely \texttt{P=[0.267 0.344 0.141 0.248]},
\texttt{Q=[0.893 0.896 0.126 0.207]} and \texttt{A=[0 1 0 1]}.}
of $\VP$ and $\VQ$.

%% file: body/impossibility.tex
\section{Strong impossibility result}

\begin{figure}
  \center
  \includegraphics[width=0.9\columnwidth]{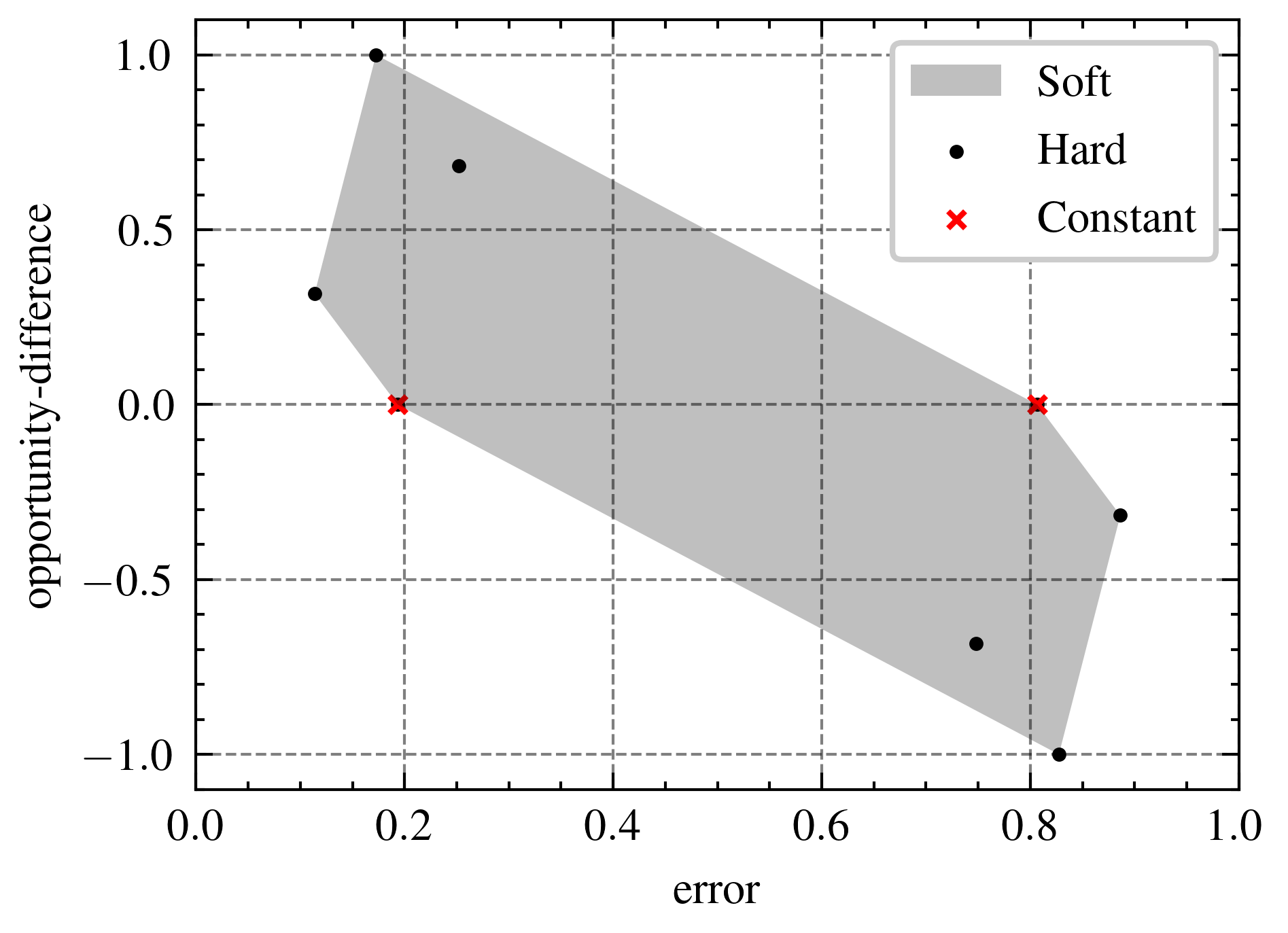}
  \caption{
    \label{fig:ex-plane}
    Randomly generated example using Algorithm~\ref{alg:plane} (Python code in supplementary material).
    The constant classifiers are vertices of the polygon, thus the constraints of equal-opportunity and non-trivial accuracy can not be satisfied simultaneously.
  }
\end{figure}

Contrasting with Figure~\ref{fig:non-example} in the previous section, Figure~\ref{fig:ex-plane} shows a data source for which the constant classifiers are vertices of the polygon.
This figure illustrates the strong incompatibility that may occur (especially in highly probabilistic distributions).
Namely, among the predictors satisfying equal-opportunity (those in the X-axis), the minimal error is achieved by a constant classifier.

In other words, there are data sources for which no predictor can achieve equal-opportunity and non-trivial accuracy simultaneously. This is Theorem~\ref{thm:plane}.

Since Theorem~\ref{thm:plane} is our strongest result, we also show how to generalize it to non-finite domains. For this purpose, and focusing on formality, we state in Definition~\ref{defn:reasonable} very precisely, for which kind of domains it applies.

\begin{definition}
  \label{defn:reasonable}
  The \emph{essential range} of a random variable $S:\Omega \to \RR^k$ is the set\[
    \{\vec s\in\RR^k: (\forall \epsilon>0)\,\, \P{\norm{S-\vec s} < \epsilon} > 0 \}
  \]
  We call a set $D\subseteq\RR^k$ an \textit{essential domain} if it is the essential range of any random variable.
\end{definition}

Definition~\ref{defn:reasonable} excludes pathological domains such as non-measurable sets, the Cantor set or the irrationals.
But it allows for isolated points, convex and closed sets, finite unions of them and countable unions of them as long as the resulting set is closed.
This includes typical domains, such as products of closed intervals $\prod_{i=1}^n [l_i, r_i]$, or the whole space $\RR^n$.

\subimport{.}{theorem-plane-short}

Finally, to conclude this section we present Example~\ref{exa:cloud}.
It shows that there are many other scenarios, not necessarily those of Theorem~\ref{thm:plane}, in which EO and non-trivial accuracy are incompatible.

\begin{longexqmple}\label{exa:cloud}
Consider a data source $(X,A,Y)$ over $\{0,1\}^3$ whose distribution is given by 
\[
\begin{array}{cccc}
x & a & \xa(x,a) & q(x,a)\\
0 & 0 & 3/8 & 9/20\\
0 & 1 & 2/8 & 15/20\\
1 & 0 & 1/8 & 15/20\\
1 & 1 & 2/8 & 16/20
\end{array}
\]

Then, (i) there are predictors satisfying equal-opportunity, (ii) there are predictors with non-trivial accuracy, but (iii) there are no predictors satisfying both. \qedhere
\end{longexqmple}

\begin{figure}
\begin{centering}
\includegraphics[width=8cm]{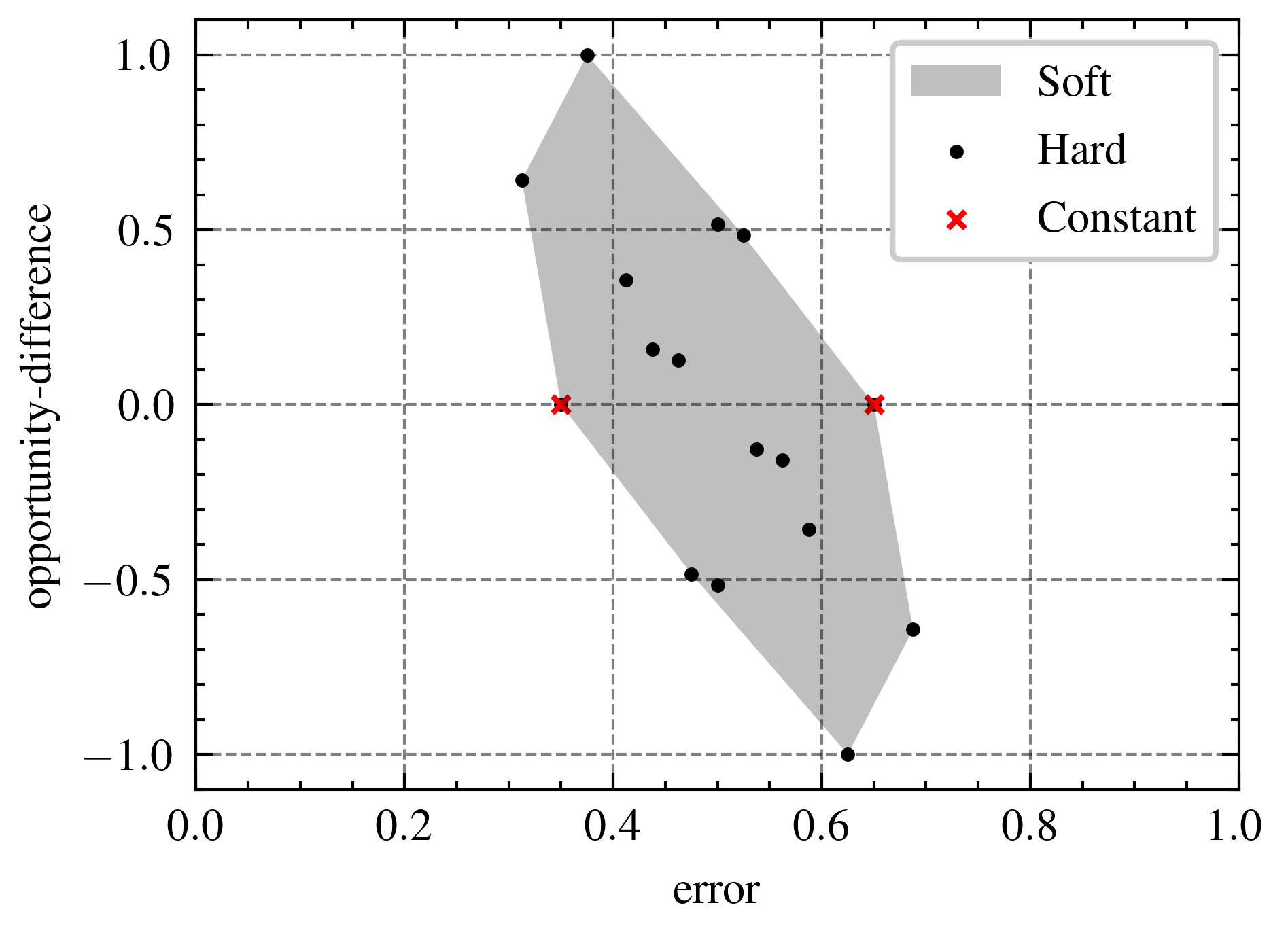}
\par\end{centering}
\caption{\label{fig:cloud}Example~\ref{exa:cloud}. One of the constant classifiers is Pareto-optimal.}
\end{figure}

Indeed, Figure~\ref{fig:cloud} depicts the region $\regionM$ for Example~\ref{exa:cloud}.
On the one hand, the set of non-trivially accurate predictors corresponds to the area with error strictly smaller than the left constant classifier.
On the other hand, the set of equal-opportunity predictors is (for this particular example) the closed segment between the two constant classifiers.
As claimed in Example~\ref{exa:cloud} (and depicted in Figure~\ref{fig:cloud}), these two sets are non-empty and do not intersect each other.

\references

%% file: body/theorem-plane-short.tex
\begin{theorem}\label{thm:plane}
For any essential domain $\calX\subseteq\RR^d$ with $|\calX|\geq 2$ there exists a data source $(X,A,Y)$ whose essential range is $\calX \times \{0,1\}^2$ and such that the accuracy $\acc{\Q}$ of any predictor $\Q\in\setQ$ that satisfies equal opportunity is at most the trivial accuracy $\tau$.
\end{theorem}

\newcommand{\LA}[0]{\ensuremath{L}}
\newcommand{\stimes}[2]{\compact{#1 \times #2}}

\begin{proof}
The complete proof is contained in the supplementary material. 
Here we highlight only the sketch, the intuition and some relevant details.

Partition the non-protected input space $\calX$ into two non-empty sets $\calX_1, \calX_2$, and the input space $\calX \times \{0,1\}$ into three regions $R_j$:
\[\begin{alignedat}{1}
  R_1 &= \calX_1 \times \{0\}\\
  R_2 &= \calX_2 \times \{0\}\\
  R_3 &= \calX \times \{1\}\\
\end{alignedat}\]

For any distribution $(\xa,q)$ for which these $3$ regions have positive probabilities, denote $\VP_j\eqdef\P{(X,A)\in R_j} >0$ and $\VQ_j\eqdef\PP{Y=1}{(X,A)\in R_j}$ for $j\in\{1,2,3\}$.
We search for constraints over $\VP$ and $\VQ$ that are feasible and cause $\acc{\Q}\leq \tau$ for any predictor $\Q\in\setQ$ satisfying EO.
As shown in the supplementary material, the following constraints suffice:
\begin{constraints}
  \item \label{enu:constr-open} $\VP\in (0,1)^3$ and, for probabilism, also $\VQ\in (0,1)^3$.
  \item \label{enu:constr-constant} The accuracy of $\Q=\hat 1$ is higher than that of $\Q=\hat 0$ (for fixing an orientation).
  \item \label{enu:constr-Bayes} $\VQ_1<\half$ and $\VQ_2,\VQ_3 > \half$.
  \item \label{enu:constr-Q} $\VQ_3+\VQ_1 \geq 1$, and
  \item \label{enu:constr-P} $\VP_1 \VQ_1 + \VP_2 \VQ_2 < \VP_3 \VQ_1$.
\end{constraints}

The last three constraints are not straightforward, but their main consequence can be explained graphically.
Let us characterize each predictor $\Q$, with a vector $\VF$ given by $\VF_j\eqdef \P{\Y=1, (X,A)\in R_j}$, so that $\Q=\hat 1$ corresponds to $\VF=\VP$.
Figure~\ref{fig:eo-plane} depicts the set of all predictors (box), and those that satisfy EO (plane).
This plane can be characterized by the two vectors $\VZ$ and $\VP$.

Constraint~\ref{enu:constr-Bayes} simply fixes the location of the Bayes classifier at $(0,\VP_2,\VP_3)$.
Constraints~\ref{enu:constr-Q} and \ref{enu:constr-P} force the gradient of the accuracy along the plane to be non-decreasing in the directions $\VZ$ and $\VP-\VZ$, so that for each predictor $\VF$ there is a pivot $\VF^*$ with higher accuracy than $\VF$ and lower accuracy than $\VP$.
As a consequence, when the constraints are satisfied, $\hat 1$ has maximal accuracy in $\setQ$.

In order to satisfy the constraints, we propose a randomized algorithm (Algorithm~\ref{alg:plane})  that generates random vectors $\VP$, $\VQ$ satisfying the five constraints, regardless of the seed and the random sampling function, e.g. uniform.
The proof is presented in the supplementary material. To corroborate, Figure~\ref{fig:ex-plane} shows a particular output of the algorithm\footnote{The output is \texttt{P=[0.131 0.096 0.772]} and \texttt{Q=[0.274 0.858 0.891]}. Also, \texttt{A=[0 0 1]} from the partition $\{R_1,R_2,R_3\}$.}.

The proof concludes by showing that given $\VP$ and $\VQ$ (produced by the algorithm), it is always possible to split $\calX\times\imA$ into the three regions $R_j$ (for $j\in\{1,2,3\}$) that satisfy $\P{(X,A)\in R_j}=\VP_j$ and $\PP{Y=1}{(X,A)\in R_j}=\VQ_j$.
\end{proof}

\begin{figure}
  \center
  \subimport{.}{3dplane-initial}
  \caption{
    \label{fig:eo-plane}
    In vectorial form, the predictors that satisfy equal-opportunity form a plane inside the rectangular box of all predictors.
  }
\end{figure}
\begin{algorithm}[H]
  \caption{Random generator for Theorem~\ref{thm:plane}.}\label{alg:plane}
  \small
  \begin{algorithmic}[1]
    \Procedure{VectorGenerator}{seed}
    \State Initialize random sampler with the seed
    \State $\VQ_1 \gets$ random in $(0,\half)$
    \State $\VQ_2 \gets$ random in $(\half,1)$
    \State $\VQ_3 \gets$ random in $(1-\VQ_1, 1)$
    \State $\VP_3 \gets$ random in $(\half, 1)$
    \State $a \gets \max\{(1-\VP_3)\VQ_1, \half - \VP_3 \VQ_3\}$
    \State $b \gets \min\{(1-\VP_3)\VQ_2, \VP_3 \VQ_1\}$
    \State $c \gets$ random in $(a, b)$
    \State $\VP_2 \gets \nicefrac{(c - \VQ_1 (1-\VP_3))}{\VQ_2 - \VQ_1}$
    \State $\VP_1 \gets 1 - \VP_3 - \VP_2$
    \State \Return $\VP, \VQ$
    \EndProcedure
  \end{algorithmic}
\end{algorithm}

%% file: body/3dplane-initial.tex
\usetikzlibrary{calc}
\usetikzlibrary{shadings}
\tdplotsetmaincoords{70}{115}
\pgfmathsetmacro{\Px}{1}
\pgfmathsetmacro{\Py}{1}
\pgfmathsetmacro{\Pz}{1}
\pgfmathsetmacro{\Zx}{0.7}

\begin{tikzpicture}[
    scale=4,
    tdplot_main_coords,
    axis/.style={->,black,thick},
    plane/.style={transparent, fill=cyan,fill opacity=0.6},
    planegrid/.style={very thin, white!80!blue},
    cubeInt/.style={draw opacity=0, fill=gray,fill opacity=.15},
    cubeFr/.style={very thin, black, draw opacity=.5},
    dashedplane/.style={very thin,dashed,black,opacity=.3},
    dashedline/.style={very thin,dashed,orange},
    vectorB/.style={->,thick,black},
    vectorIn/.style={->,thick,black!10!blue},
    vector/.style={->,thick,black!10!orange, text=black},
  ]
  \coordinate (O) at (0,0,0);
  \coordinate (x) at (\Px,0,0);
  \coordinate (y) at (0,\Py,0);
  \coordinate (z) at (0,0,\Pz);
  \coordinate (Z) at (\Zx,\Py,0);
  \coordinate (P) at ($(x) + (y) + (z)$);
  \coordinate (F) at ($0.5*(P) - 0.25*(Z)$);
  \coordinate (FF) at ($(F) + 0.75*(Z)$);
  \coordinate (Bayes) at ($(x) + (y)$);

	\draw[axis,tdplot_main_coords] (O) -- ($1.2*(x)$) node[anchor=east]{$\VF_3$};
	\draw[axis,tdplot_main_coords] (O) -- ($1.2*(y)$) node[anchor=west]{$\VF_2$};
	\draw[axis,tdplot_main_coords] (O) -- ($1.1*(z)$) node[anchor=east]{$\VF_1$};
	
	\draw[plane,
    shade, left color=blue, right color=green, 
  ] (O) -- (Z) -- (P) -- ($(P) - (Z)$) -- cycle;

  \coordinate (gridI) at ($0.33*(P)-0.33*(Z)$);
  \coordinate (gridII) at ($0.33*(Z)$);
  
  \draw[cubeInt] (z) -- ($(z) + (x)$) -- (P) -- ($(z) + (y)$) -- cycle;
	\draw[cubeInt] (x) -- ($(x) + (y)$) -- (P) -- ($(x) + (z)$) -- cycle;
	\draw[cubeInt] (y) -- ($(y) + (z)$) -- (P) -- ($(y) + (x)$) -- cycle;
  
  \draw[vector] (0,0,0) -- (F) node[anchor=south]{$\VF$};
	\draw[vector] (0,0,0) -- (FF) node[anchor=west]{$\VF^*$};
	\draw[dashedline] (F) -- (FF) node[anchor=west]{};

  \draw[cubeFr] ($(x) + (y)$) -- (x) -- ($(x) + (z)$) -- (P);
  \draw[cubeFr] ($(y) + (z)$) -- (y) -- ($(y) + (x)$) -- (P);
  \draw[cubeFr] ($(z) + (x)$) -- (z) -- ($(z) + (y)$) -- (P);

	\draw[vectorB] (0,0,0) -- (P) node[anchor=west]{$\VP$};
	\draw[vectorB] (0,0,0) -- (Z) node[anchor=west]{$\vec Z$};
	\draw[vectorB] (0,0,0) -- (Bayes) node[anchor=north]{Bayes};
	
  \filldraw[vectorB] (O) circle (0.3pt) node[anchor=south east] {$\vec 0$};

\end{tikzpicture}

%% file: body/probabilism.tex
\section{Probabilistic vs deterministic sources}

In this section we compare the tension between error and opportunity-difference when the data source is deterministic and probabilistic. 
Particularly, we show that some known properties that apply for the discrete case may fail to hold for the probabilistic one, and under what conditions this happens.

As commented in the introduction, analyzing the more general probabilistic case is motivated by two different reasons.
On the one hand, scientifically, it provides a more general perspective.
On the other hand, ethically, it enables a more realistic representation of a fair environment: one in which the information in $(X,A)$ may be insufficient to conclude the correct decision $Y$; in which $(X,A)$ are prone to errors; or in which real-life constraints force the correct decision to be different for identical inputs $(x,a)$, e.g. two identical candidates for a single job position.

\subsection{Deterministic sources}

Under the assumption that the data source is deterministic, there are some important existing results showing the compatibility between equal-opportunity and high accuracy:

\begin{fact}
\label{enu:d1}
Assuming a deterministic data source, if $\tau<1$, then there is always a non-trivial predictor, for instance the Bayes classifier $\thr{Q}$. Otherwise (degenerated case) all predictors are trivially accurate.
\end{fact}

\begin{fact}
\label{enu:d2}
Assuming a deterministic data source, the Bayes classifier $\thr{Q}$ satisfies equal-opportunity necessarily.
\end{fact}

As a consequence, EO and maximal accuracy (thus also non-trivial accuracy) are always compatible provided $\tau<1$, because the Bayes classifier satisfies both.
This is a celebrated fact and it was part of the motivations of \cite{hardt2016equality} for defining equal-opportunity, because other notions of fairness, including statistical parity, are incompatible with accuracy.

\subsection{Probabilistic sources}

If we allow the data source to be probabilistic, the results of the deterministic case change.
In particular, Fact~\ref{enu:d1} is generalized by Proposition~\ref{prop:NTA} and Fact~\ref{enu:d2} is affected by Proposition~\ref{prop:B-EO} and Example~\ref{exa:cloud}.

Analogous to $\tau$ for deterministic sources, we define a secondary reference value $\tau^{*}\in[0,1]$. We let \[
  \tau^{*}\eqdef\max\left\{ \P{Q\geq\half},\,\P{Q\leq\half}\right\},
\]
highlighting that (i) $Q=q(X,A)$ is a random variable varying in $[0,1]$, (ii) $\tau$ and $\tau^*$ are equal when the data source is deterministic, and (iii) the condition $\tau=1$ implies $\tau^*=1$, but not necessarily the opposite.

As shown in Proposition~\ref{prop:NTA}, the equation $\tau^*=1$ characterizes the necessary and sufficient conditions on the data source for non-trivially accurate predictors to exist.

Particularly, in the deterministic case, we have $\tau^{*}=\tau$, and Proposition~\ref{prop:NTA} resembles Fact~\ref{enu:d1}.

\begin{proposition}
\label{prop:NTA}
(Characterization of the impossibility of non-trivial accuracy)

For any arbitrary source distribution $(\xa,q)$, non-trivial predictors exist
if and only if $\tau^{*}<1$.
\end{proposition}
\begin{proof}
The proof is in the supplementary material. Intuitively, if $\P{Q\geq\half}=1$, then predicting $1$ for any input is optimal, and vice versa.
\end{proof}

Finally, in Proposition~\ref{prop:B-EO} and its proof, we show a simple family of probabilistic examples for which equal-opportunity and optimal accuracy (obtained by the Bayes classifier) are not compatible.
This issue does not merely arise from the fact that the Bayes classifier is hard while the data distribution is soft.
Adding randomness to the classifier does not solve the issue.
To justify this, and also for completeness, we considered the soft predictor $Q$ and showed that it also fails to satisfy equal-opportunity.

\subimport{.}{theorem-bayes-eo}

As a remark, notice that the data sources proposed in the proof of Proposition~\ref{prop:B-EO}, contrast the extreme case $Y=A$ because they allow some mutual information between $X$ and $Y$ after $A$ is known, as one would expect in a real-life distribution.
Nevertheless, there is an evident inherent demographic disparity in these distributions, and this can be the reason why equal-opportunity hinders optimal accuracy for these examples.

\references

%% file: body/theorem-bayes-eo.tex
\begin{proposition}
\label{prop:B-EO}

There are data sources for which neither the Bayes classifier $\thr{Q}$ nor the predictor $Q$ satisfy equal-opportunity.

\end{proposition}

\begin{proof}

Fix any data source with $\P{A=a,Y=1}>0$ for each $a\in\imA$, pick an arbitrary ($(X,A)$-measurable) function $c:\imX\to(0,\half)$ and let

\[
q\left(x,a\right)\eqdef\begin{cases}
\half-c(x) & \text{ if }a=0\\
\half+c(x) & \text{ if }a=1
\end{cases}
\]
for each $(x,a)\in\imXA$.

\def\tmpE#1#2{\EE{#1}{A=#2,Y=1}}


Since we know that $\thr{Q}(x,a)=a$, then the term $\tmpE{\thr{Q}(X,A)}{a}$ can be reduced more simply into $\tmpE{A}{a}=a$. Therefore, the Bayes classifier satisfies $\oppDiff{\thr{Q}}=1-0>0$.

Regarding $Q$, we have $\tmpE{Q}{1}=\half+\tmpE{c(X)}{1}$ and $\tmpE{Q}{0}=\half+\tmpE{c(X)}{0}$.
Notice from the range of $c$, that $\tmpE{Q}{1}\in(\half,1)$ and $\tmpE{Q}{0}\in(0,\half)$.
Hence $\oppDiff{Q}>0$.

Therefore neither $\thr{Q}$ nor $Q$ satisfy equal-opportunity.
  
\end{proof}

%% file: body/characterization/characterization.tex
\def\EP#1#2{\P{#2}\EE{#1}{#2}}%

\def\q#1{\hat q_{#1}}%
\def\a#1{\alpha_{#1}}%

\section{Sufficiency condition}

In this section, we provide a simple sufficient (but not necessary) condition (Theorem~\ref{thm:sufficient}) that guarantees that equal-opportunity and non-triviality are compatible.
It is not very restrictive and it is valid for discrete, continuous and mixed data sources.
Therefore, it may be used as a minimal assumption for any research work on equal-opportunity dealing with probabilistic data sources.
It can also be used to verify whether a data source $(X,A,Y)$ of a particular application is pathogenic for equal-opportunity or not.

Figure~\ref{fig:QAregions} summarizes  the sufficiency condition in simple manner.
The proof consists of showing that when the 4 events highlighted in Figure~\ref{fig:QAregions} have positive probabilities, then it is possible to use one of them to improve the performance of the best constant classifier and another one to compensate for equal opportunity.

\begin{figure}[ht]
  \center
  \subimport{.}{figure2}
  \caption{
    \label{fig:QAregions}
    Sufficiency condition: If the 4 blue events have positive probability, then equal-opportunity and non-triviality are compatible.
  }
\end{figure}

\begin{lemma}\label{lemma:EQQ}
Assume, for EO to be well defined, that $\P{Y=1,A=a}>0$ for each $a\in\imA$. For any predictor $\Q$, we have
\[
    \PP{\hat Y=1}{Y=1,A=a} =
    \frac{\EE{\Q Q}{A=a}}{\EE{Q}{A=a}}
\]
\end{lemma}
\begin{proof}
Proved in the supplementary material.
\end{proof}

\begin{theorem}\label{thm:sufficient}
  (Sufficiency condition, Figure~\ref{fig:QAregions})

  For any given data source $(X,A,Y)$, not-necessarily discrete, if
  \[
    \P{Q>\half,A=a}, \P{Q<\half,A=a}>0
  \]
  for each $a\in\set{0,1}$, then equal-opportunity and non-triviality are compatible.
\end{theorem}

\begin{proof}
We begin by noticing that $\P{Q>\half, A=a}>0$ implies $\P{Y=1, A=a}>0$ for each $a\in\set{0,1}$, thus equal-opportunity is well-defined.

The proof is divided into two cases depending on which constant classifier is optimal (either $\hat 0$ or $\hat 1$).
The distinction is needed because equal-opportunity treats $Y=1$ and $Y=0$ differently.

\proofcase{Assume $\err{\hat 0}\leq \err{\hat 1}$.}

We will show that there are constants $\q0, \q1 \in [0,1]$ such that the following predictor satisfies equal-opportunity and non-triviality.
\[
  \Q \eqdef
  \begin{cases}
    0 & \text{if } Q\leq\half\\
    \q0 & \text{if } Q>\half, A=0\\
    \q1 & \text{if } Q>\half, A=1\\
  \end{cases}
\]

For $\Q$ to satisfy non-triviality, it suffices to have smaller error than $\hat 0$.
This holds whenever $\q0$ or $\q1$ are positive, because for each $a\in\set{0,1}$, the optimal classification over the region $\set{Q>\half, A=a}$ is $1$, thus any positive value $\q{a}$ improves the misclassification of $\hat 0$.
For this reason, we restrict $\q0, \q1 \in (0,1]$.

Regarding equal-opportunity, recall from Lemma~\ref{lemma:EQQ} that 
\[
   \PP{\hat Y}{Y=1,A=a} =
   \frac{\EE{\Q Q}{A=a}}{\EE{Q}{A=a}}
\]
Let us call $\a{a}\eqdef \EE{\Q Q}{A=a}$ to the numerator.
Since $\Q=0$ for $Q\leq\half$, then $\a{a}$ may be computed as
\[
  \a{a}=\q{a}\EE{Q}{A=a,Q>\half}\PP{Q>\half}{A=a}
\]
and it is positive.

Hence, equal-opportunity may be stated as
\[
   \frac{\q1}{\q0} =
   \frac%
   {\a0 \EE{Q}{A=1}}%
   {\a1 \EE{Q}{A=0}}
\]
The right hand side term is always well-defined, and it is a positive real number.
Since $\q0$ and $\q1$ can be made arbitrarily small, there are always solutions to this equation in the range $\q0, \q1 \in (0,1]$.

\proofcase{Assume $\err{\hat 1} < \err{\hat 0}$.}

Analogously, for $\q0, \q1 \in [0,1]$, let $\Q$ be given by
\[
  \Q \eqdef
  \begin{cases}
    1 & \text{if } Q\geq\half\\
    \q0 & \text{if } Q<\half, A=0\\
    \q1 & \text{if } Q<\half, A=1\\
  \end{cases}
\]

If $\q0,\q1<1$, then $\Q$ satisfies non-triviality, because it has less error than $\hat{1}$.
Hence, we restrict $\q0\q1 < 1$.

Equal-opportunity, may be equivalently stated in terms of $\PP{\hat Y=0}{Y=1,A=a}$ because it is the complement of $\PP{\hat Y=1}{Y=1,A=a}$.
Recall from Lemma~\ref{lemma:EQQ} that 
\[
  \PP{\hat Y=0}{Y=1,A=a} =
  \frac{\EE{(1-\Q) Q}{A=a}}{\EE{Q}{A=a}}
\]
Let us call $\a{a}\eqdef \EE{(1-\Q) Q}{A=a}$ to the numerator.
Since $(1-\Q)=0$ for $Q\geq\half$, then $\a{a}$ may be computed as
\[
  \a{a}=(1-\q{a})\EE{Q}{A=a,Q<\half}\PP{Q<\half}{A=a}
\]
and it is non-negative.

Hence, equal-opportunity my be stated as
\[
   (1-\q1) \a1 \EE{Q}{A=0}=
   (1-\q0) \a0 \EE{Q}{A=1}
\]
If $\a1 \EE{Q}{A=0}=0$, we let $\q0=1$ and $\q1=\half$.
If $\a0 \EE{Q}{A=1}=0$, we let $\q1=1$ and $\q0=\half$.
And if none of the two is zero, we use the same argument as in the first case:
since $1-\q0$ and $1-\q1$ can be made arbitrarily small, there are always solutions to this equation in the range $\q0, \q1 \in [0,1)$.
\end{proof}

\references

%% file: body/characterization/figure2.tex
\begin{tikzpicture}
  
  \coordinate (LB) at (0,0.0);
  \coordinate (LT) at (0,3.0);
  \coordinate (RB) at (6,0.0);
  \coordinate (RT) at (6,3.0);

  \coordinate (LM) at (0,1.5);
  \coordinate (RM) at (6,1.5);
  \coordinate (lB) at (2,0.0);
  \coordinate (lT) at (2,3.0);
  \coordinate (rB) at (4,0.0);
  \coordinate (rT) at (4,3.0);
  
  \draw[line width=0.8pt, fill=cyan!20]
    (LB) --
    (LT) --
    (RT) --
    (RB) --
    cycle;

  \draw[line width=0.8pt, fill=white]
    (lB) to[out=80, in=-80]
    (lT) --
    (rT) to[out=-100, in=100]
    (rB) --
    cycle;
  
    \draw[line width=0.8pt]
      (LM) to[out=-5, in=175] (RM);
  
  \node[align=left] at (1,0.8) {\shortstack{$Q<\half$\\$A=0$}};
  \node[align=left] at (1,2.1) {\shortstack{$Q<\half$\\$A=1$}};

  \node[align=left] at (3,0.8) {\shortstack{$Q=\half$\\$A=0$}};
  \node[align=left] at (3,2.1) {\shortstack{$Q=\half$\\$A=1$}};

  \node[align=left] at (5,0.8) {\shortstack{$Q>\half$\\$A=0$}};
  \node[align=left] at (5,2.1) {\shortstack{$Q>\half$\\$A=1$}};
\end{tikzpicture}

%% file: body/conclusion.tex
\section{Conclusion and future work}

Our work extends existing results about equal-opportunity and accuracy from a deterministic data source to a probabilistic one.
The main result, Theorem~\ref{thm:plane}, states that for certain probabilistic data sources, no predictor can achieve equal-opportunity and non-trivial accuracy simultaneously.
We also provided a sufficient condition on the data source under which EO and non-trivial accuracy are guaranteed to be compatible.

Our method focuses on the fairness notion of equal-opportunity, which seeks for equal true positive rates.
A symmetric analysis can be carried out for equal false positive rates using the same ideas we provided.
Since the notion of equal-odds seeks for both equal true positive rates and equal false positive rates, our methodology and results can be used for equal-odds.
However, our method is not compatible with statistical parity or with fairness notions based on individual fairness.

An interesting question left for future work is whether the scenarios in which equal-opportunity and non-trivial accuracy are incompatible require the data source to be unfair on its own in some sense.
If this is true, it would provide additional justification for equal-opportunity as a fairness notion.

Other future lines of research include  characterizing completely the conditions under which equal-opportunity and non-trivial accuracy are compatible, studying the trade-off and  Pareto-optimality between accuracy and  opportunity-difference, and bounding the opportunity-difference by taking into account  the learning process and the statistical sampling.
In addition, we plan to measure the accuracy gap between the Bayes classifier and the most accurate predictor that satisfies equal-opportunity.

\references

%% file: body/polygon.tex
\newcommand{\centerclaim}[1]{%
  \begin{changemargin}{0.1\columnwidth}{0.1\columnwidth}\begin{center}%
    \textit{#1}%
  \end{center}\end{changemargin}%
}

\section{The error vs opportunity-difference region}

\begin{figure}
  \center
  \includegraphics[width=0.9\columnwidth]{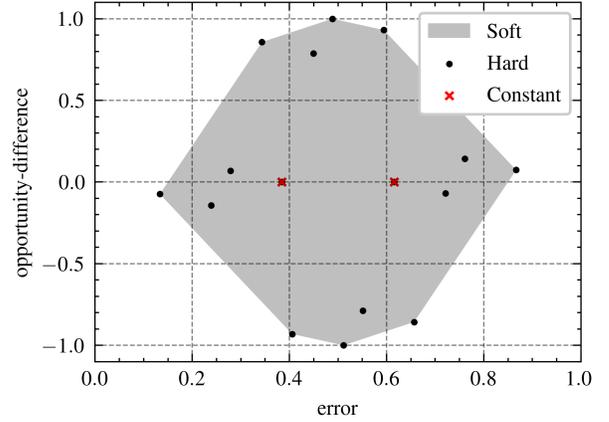}
  \caption{
    \label{fig:non-example}
    Region $\regionM$ for an arbitrary source distribution.
    The data source is not deterministic because the Bayes classifier does not satisfy equal-opportunity.
  }
\end{figure}

In this section, we analyze the region $\regionM\subseteq [0,1]\times[-1,+1]$ given by \[
  \regionM \eqdef \{(\err{\Q},\oppDiff{\Q}): \Q\in\setQ\}
\]
which represents the feasible combinations of the evaluation metrics (error and opportunity-difference) that can be obtained for a given source distribution $(\xa, q)$. This region determines the tension between error and opportunity difference.
Figure~\ref{fig:non-example} shows an example of the region $\regionM$.

The results presented in this section assume that the data source is discrete and its range is finite.
We will use the following vectorial notation to represent both the distribution $(\xa, q)$ and any arbitrary predictor $\Q\in\setQ$.

\begin{longdefinition}\label{def:vector-setup}
    Suppose $(X,A)$ can only take a finite number of outcomes $\{(x_i,a_i)\}_{i=1}^n$ (each with positive probability) for some integer $n>0$.
    In order to represent $\xa$, $q$ and any $\Q\in\setQ$ respectively, let $\VP,\VQ,\VF\in\RR^n$ be the vectors given by\[
    \begin{alignedat}{1}
      \VP_i &\eqdef \P{X=x_i, A=a_i} \\
      \VQ_i &\eqdef \PP{Y=1}{X=x_i, A=a_i} \\
      \VF_i &\eqdef \P{\Y=1, X=x_i, A=a_i} \\
    \end{alignedat}
    \]
    For notation purposes, let also $\VQ^{(0)},\VQ^{(1)}\in\RR^n$ be given by $\VQ^{(a)}_i \eqdef \VQ_i \cdot \one{a_i=a}$, and, following the definition of $\err{\Q}$ and $\oppDiff{\Q}$, let\[
    \begin{alignedat}{1}
      \err{\VF} &\eqdef \inner{\VP}{\VQ} + \inner{\VF}{1-2\VQ} \\
      \oppDiff{\VF} &\eqdef
        \frac{\inner{\VF}{\VQ^{(1)}}}{\inner{\VP}{\VQ^{(1)}}}
        - \frac{\inner{\VF}{\VQ^{(0)}}}{\inner{\VP}{\VQ^{(0)}}} \\
    \end{alignedat}
    \] where $\inner{\,\cdot\,}{\,\cdot\,}$ denotes the inner product:\[
      \inner{u}{v}\eqdef u_1 v_1 +\cdots+ u_n v_n.\qedhere
    \]
\end{longdefinition}

Regarding Definition~\ref{def:vector-setup}, we highlight three important remarks:
\begin{enumerate}
  \item $\VQ\in [0,1]^n$, $\VP\in(0,1]^n$, $\norm{\VP}_1=1$ and $\VF$ lies in the rectangular $n$-dimensional box given by \[
    0 \vleq \VF \vleq \VP,
  \]where $\vleq$ denotes the componentwise order in $\RR^n$, i.e. $0 \leq \VF_i\leq \VP_i$ for each  $i\in\{1, ..., n\}$. Moreover, from the definition of $\VP$ and $\VF$, the vertices of this rectangular box correspond precisely with the deterministic predictors.
  \item The vectorial definitions of error and opportunity-difference correspond to those of the non-vectorial case as shown in Lemma~\ref{lem:vector-metrics}. Moreover, their gradients are constant. 
  \item There is a one-to-one correspondence between the predictors $\q\in\setQ$ and the vectors $\VF$ that satisfy $0 \vleq \VF \vleq \VP$. Indeed, each predictor is uniquely given by its pointwise values $\q(x_i, a_i) = \frac{\VF_i}{\VP_i}$ and each vector by its pointwise coordinates $\VF_i = \VP_i \q(x_i, a_i)$. Therefore\[
    \regionM = \{(\err{\VF},\oppDiff{\VF}): 0 \vleq \VF \vleq \VP\}
  \]
\end{enumerate}

We now make use of results from a different research area in mathematics, geometry, to conclude the main properties of the region $\regionM$.

\begin{theorem}
\label{thm:polygon}
Assuming a discrete data source with finitely many possible outcomes, the region $\regionM$ of feasible combinations of error versus opportunity-difference satisfies the following claims.
\begin{enumerate}
  \item $\regionM$ is a convex polygon.
  \item The vertices of the polygon $\regionM$ correspond to some deterministic predictors.
  \item $\regionM$ is symmetric with respect to the point $(\half, 0)$.
\end{enumerate}
\end{theorem}
\begin{proof}
\setcounter{proofpart}{0}

Assume the notation of Definition~\ref{def:vector-setup}.

\proofpart{} In geometrical terms, $\regionM$ is the result of applying an affine transformation, i.e. a linear transformation and a translation, to the $n$-dimensional \emph{polytope} given by $0 \vleq \VF \vleq \VP$.

Affine transformations are known to map polytopes into polytopes (See Chapter 3 of \cite{grunbaum2013convex}), therefore $\regionM$ must be a $2$-dimensional polytope, i.e. the region $\regionM$ is a convex polygon. In theory, this region may also be a $1$-dimensional segment, but this can only occur in the extreme case that $Q=\half$ (with probability 1).

\proofpart{} The vertices of a polytope, also called extremal points, are the points in the polytope that are not in the segment between any two other points in the polytope.
It is known from geometry theory that affine mappings preserve collinearity, i.e. they map segments into segments,
thus they map non-vertices into non-vertices.
As a consequence, the vertices of the polygon $\regionM$ correspond to some vertices vertices of the polytope $0 \vleq \VF \vleq \VP$, that is, to some deterministic classifiers.

\proofpart{} Notice (Lemma~\ref{lem:vector-symmetry}) that \[
\begin{alignedat}{1}
  \err{\VP-\VF} &= 1 - \err{\VF} \\
  \oppDiff{\VP-\VF} &= - \oppDiff{\VF} \\
\end{alignedat}
\]
This implies that for each point $(\err{\VF},\oppDiff{\VF})\in\regionM$, there is another one, namely $(\err{\VP-\VF},\oppDiff{\VP-\VF})\in\regionM$ that is symmetrical to the former w.r.t the point $(\half, 0)$.
Geometrically, this means that the polygon $\regionM$ is symmetric with respect to the point $(\half, 0)$.
\end{proof}

The reader is invited to visualize the aforementioned properties of $\regionM$ in Figure~\ref{fig:non-example}, which depicts the region $\regionM$ for a particular instance\footnote{Namely \texttt{P=[0.267 0.344 0.141 0.248]},
\texttt{Q=[0.893 0.896 0.126 0.207]} and \texttt{A=[0 1 0 1]}} of $\VP$ and $\VQ$.

%% file: body/theorem-plane.tex
\begin{theorem}\label{thm:plane}
For any essential domain $\calX\subseteq\RR^d$ with $|\calX|\geq 2$ there exists a data source $(X,A,Y)$ whose essential range is $\calX \times \{0,1\}^2$ and such that the accuracy $\acc{\Q}$ of any predictor $\Q\in\setQ$ that satisfies equal opportunity is at most the trivial accuracy $\tau$.
\end{theorem}

\begin{proof}
\newcommand{\LA}[0]{\ensuremath{L}}
\newcommand{\stimes}[2]{\compact{#1 \times #2}}
\setcounter{proofpart}{0}

The proof is divided in four parts. We will (i) reduce the problem into an algebraic one; (ii) find the linear constraints that solve the algebraic problem when satisfied; (iii) provide an algorithm that generates vectors that satisfy the linear constraints; and finally, (iv) convert the vectorial solution back into a distribution $(\xa,q)$ for the given domain.

\proofpart{Reduction to an algebraic problem.}

Partition the non-protected input space $\calX$ into two non-empty sets $\calX_1, \calX_2$, and the input space $\calX \times \{0,1\}$ into three regions $R_j$:
\[\begin{alignedat}{1}
  R_1 &= \calX_1 \times \{0\}\\
  R_2 &= \calX_2 \times \{0\}\\
  R_3 &= \calX \times \{1\}\\
\end{alignedat}\]

For any distribution $(\xa,q)$ for which these $3$ regions have positive probabilities, denote $\VP_j\eqdef\P{(X,A)\in R_j} >0$ and $\VQ_j\eqdef\PP{Y=1}{(X,A)\in R_j}$ for $j\in\{1,2,3\}$. 
We search for constraints over $\VP$ and $\VQ$ that are feasible and cause $\acc{\Q}\leq \tau$ for any fair predictor $\Q\in\setQ$ satisfying EO.
The first such constraint is
\begin{constraints}
  \item \label{enu:constr-open} $\VP,\VQ\in (0,1)^3$.
\end{constraints}
That is, we require $\VP_j$ to be positive, and $Y$ to have at least some degree of randomness in each region.

Given a reference predictor $\Q$, let $\VF\in [0,1]^3$ be the vector given by $\VF_j\eqdef \P{\Y=1, (X,A)\in R_j}$. Lemma~\ref{lem:vector-metrics} shows that the accuracy and the opportunity-difference of any predictor $\Q$ can be computed from $\VP$, $\VQ$ and $\VF$ as
\[\begin{alignedat}{1}
  \acc{\Q} & = \inner{\VF}{2\VQ-1} + C_{\VQ} \\
  \oppDiff{\Q} & = \frac{\VF_3}{\VP_3} - \frac{\VF_1 \VQ_1 + \VF_2 \VQ_2}{\VP_1 \VQ_1 + \VP_2 \VQ_2} \\
\end{alignedat}\] 

where $C_{\VQ}\eqdef 1 - \inner{\VP}{\VQ}$ is a constant and the operator $\inner{\cdot}{\cdot}$ denotes the inner product explained in Definition~\ref{def:vector-setup}.
Since we are interested in relative accuracies with respect to the trivial predictors, the constant $C_{\VQ}$ is mostly irrelevant.
For this reason, we let $\LA(\VF)\in[-1,1]$ denote the non-constant component of the accuracy $\LA(\VF) \eqdef \inner{\VF}{2\VQ-1}$.

Both accuracy and opportunity-difference are completely determined for any predictor by the vectors $\VP$, $\VQ$ and $\VF$ as shown above. Moreover, both quantities are linear with respect to $\VF$.

\begin{figure}
  \center
  \subimport{.}{3dplane-initial}
  \caption{
    \label{fig:eo-plane}
    In vectorial form, the predictors that satisfy equal-opportunity form a plane inside the rectangular box of all predictors.
  }
\end{figure}
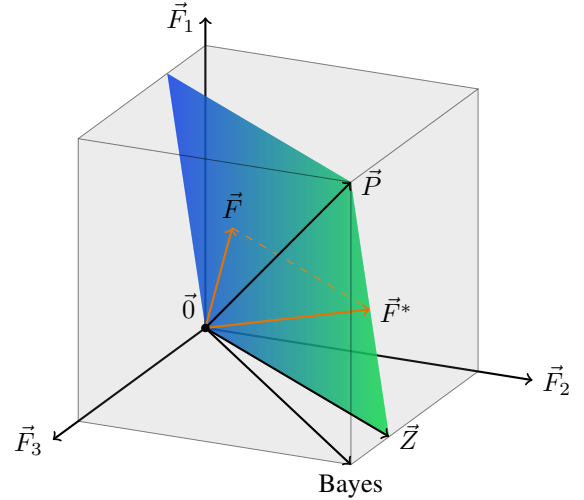

Regarding equal-opportunity, the constraint $\oppDiff{\Q}=0$ forms a plane in $\RR^3$, depicted in Figure~\ref{fig:eo-plane}. This plane passes through the origin, is determined by $\VP$ and $\VQ$, and contains all vectors $\VF$ (restricted to $0 \leq \VF_j \leq \VP_j$) that satisfy
\[
  \VF_3 (\VP_1 \VQ_1 + \VP_2 \VQ_2) - \VP_3(\VF_1 \VQ_1 + \VF_2 \VQ_2)=0
\]
or equivalently, all vectors $F$ that are normal to the vector $\compact{(-\VP_3 \VQ_1, -\VP_3 \VQ_2, \VP_1 \VQ_1+\VP_2 \VQ_2)}$.

Regarding accuracy, the two constant predictors correspond to $\VF=\vec 0$ and $\VF=\VP$, thus $\tau = C_Q + \max\{\LA(\vec 0), \LA(\VP)\}$. Importantly, both of them lie on the equal-opportunity plane.

The problem is now reduced to finding vectors $\VP$ and $\VQ$ such that all vectors $\VF$ in the equal-opportunity plane satisfy $\LA(\VF) \leq \max\{\LA(\vec 0), \LA(\VP)\}$.

\proofpart{Constraints for the algebraic solution.}

To fix an orientation, let us impose these constraints:
\begin{constraints}
  \item \label{enu:constr-constant} Among the constant predictors, the accuracy of $\VF=\VP$ is higher than that of $\VF=\vec 0$.
  This is $L(\VP) > 0 = L(\vec 0)$.
  \item \label{enu:constr-Bayes} The Bayes classifier is located at $(0,\VP_2,\VP_3)$ as in Figure~\ref{fig:eo-plane}. Algebraically this means $\VQ_1<\half$ and $\VQ_2,\VQ_3 > \half$.
\end{constraints}

In order to derive the constraints that make the scalar field $L$ maximal at $\VP$ over the plane, consider the vector $\VZ$ that lies on the plane and has minimal $\VZ_1$ and maximal $\VZ_2$, i.e.
\[
  \VZ \eqdef (0,\VP_2, \VP_3 \frac{\VP_2 \VQ_2}{\VP_1 \VQ_1+\VP_2 \VQ_2})
\]
Since the gradient of $L$ is given by $2\VQ-1$ and has signs $(-,+,+)$, then for any vector $\VF$ in the plane, there is $\VF^*$ in the segment between $\VP$ and $\VZ$ such that $\VF_1=\VF^*_1$ and $\LA(\VF^*) \geq \LA(\VF)$ (refer to Figure~\ref{fig:eo-plane}).
This implies that the $L$ attains its maximal value on the segment between $\VP$ and $\VZ$.
Hence, for $L$ to be maximal at $\VP$, it would suffice to have $\LA(\VP) > \LA(\VZ)$.
As shown in Lemma~\ref{lem:plane1}, this can be achieved by imposing, in addition,
\begin{constraints}
  \item \label{enu:constr-Q} $\VQ_3+\VQ_1 \geq 1$, and
  \item \label{enu:constr-P} $\VP_1 \VQ_1 + \VP_2 \VQ_2 < \VP_3 \VQ_1$.
\end{constraints}

\proofpart{Solution to the constraints.}\label{sbuproof:sol}

Algorithm~\ref{alg:plane} is a randomized algorithm that generates random vectors.
We will prove that the output vectors $\VP$ and $\VQ$ satisfy the constraints of the previous parts of this proof, regardless of the seed and the random sampling function, e.g. uniform.

\begin{algorithm}[H]
  \caption{Random generator for Theorem~\ref{thm:plane}.}\label{alg:plane}
  \small
  \begin{algorithmic}[1]
    \Procedure{VectorGenerator}{seed}
    \State Initialize random sampler with the seed
    \State $\VQ_1 \gets$ random in $(0,\half)$
    \State $\VQ_2 \gets$ random in $(\half,1)$
    \State $\VQ_3 \gets$ random in $(1-\VQ_1, 1)$
    \State $\VP_3 \gets$ random in $(\half, 1)$
    \State $a \gets \max\{(1-\VP_3)\VQ_1, \half - \VP_3 \VQ_3\}$
    \State $b \gets \min\{(1-\VP_3)\VQ_2, \VP_3 \VQ_1\}$
    \State $c \gets$ random in $(a, b)$
    \State $\VP_2 \gets \nicefrac{(c - \VQ_1 (1-\VP_3))}{\VQ_2 - \VQ_1}$
    \State $\VP_1 \gets 1 - \VP_3 - \VP_2$
    \State \Return $\VP, \VQ$
    \EndProcedure
  \end{algorithmic}
\end{algorithm}

Two immediate observations about Algorithm~\ref{alg:plane} are that the construction of $\VQ$ implies that constraints~\ref{enu:constr-Bayes} and \ref{enu:constr-Q} are satisfied, and the construction of $\VP$ implies $\VP_1+\VP_2+\VP_3=1$. To prove the correctness of the algorithm, it remains to prove that (i) $a < b$ (otherwise the algorithm would not be well-defined), that (ii) $\VP_2\in(0,1)$ for constraint~\ref{enu:constr-open}, and also that (iii) constraints \ref{enu:constr-constant} and \ref{enu:constr-P} are satisfied.
For better readability, the algebraic proof of these claims are moved to Lemma~\ref{lem:plane2}.

\proofpart{Construction of the distribution.}

Generate a pair of vectors $\VP$ and $\VQ$ using the algorithm of the previous part (Part~\ref{sbuproof:sol}).
The first goal is to partition $\calX$ into $\calX_1$ and $\calX_2$ to generate the regions $R_1,R_2$ and $R_3$.
The second goal is to define $\xa$ in such a way that $\P{(X,A)\in R_j}=\VP_j$ for each $j\in\{1,2,3\}$.
The third and last goal is to define $q$ so that $\EE{Q}{(X,A)\in R_j} = \VQ_j$ for each $j$.
This can be done immediately by letting $q(x,a)\eqdef\VQ_j$ for all $(x,a)\in R_j$.
Thus only the first two goals remain.

For the first goal, since $|\calX|\geq 2$, we may create a simple Voronoi clustering diagram by choosing two different arbitrary points $s_1, s_2\in\calX$, and letting $\compact{\calX_1\eqdef\{s\in\calX : \norm{s-s_1}\leq \norm{s-s_2}\}}$ and  $\calX_2\eqdef\calX\setminus\calX_1$.

For the second goal, since $\calX$ is an essential domain, there exists a random variable $S$ whose essential range is $\calX$.
Notice that $\P{S\in \calX_j} \geq \P{\norm{S-s_j}< \nicefrac{\norm{s_1-s_2}}{2}}>0$ for each $j\in\{1,2\}$.
For each ($(X,A)$-measurable) event $E$, let $E_a \eqdef \{x: (x,a)\in E\}$, and define $\xa(E)$ as
\[\begin{alignedat}{1}
  \P{(X,A)\in E} &\eqdef \sum_{a=0,1} \P{X \in E_a, A=a}\\
  \P{X\in E_0, A=0} &\eqdef \sum_{j=1,2} \PP{S\in E_0}{S\in\calX_j} \VP_j\\
  \P{X\in E_1, A=1} &\eqdef \P{S\in E_1}\VP_3\\
\end{alignedat}\]
This forces $\P{(X,A)\in R_j}=\VP_j$ for each $j\in\{1,2,3\}$ as desired.
\end{proof}

%% file: body/theorem-nta.tex
\begin{proposition}
\label{prop:NTA}

(Characterization of the impossibility of non-trivial accuracy)

For any arbitrary source distribution $(\xa,q)$, non-trivial predictors exist
if and only if $\tau^{*}<1$.

\end{proposition}

\begin{proof}

We will prove that all predictors are trivially accurate if and only if $\tau^{*}=1$.

$(\Leftarrow)$ Suppose $\tau^{*}=1$, i.e. $\P{Q\leq\half}=1$ or $\P{Q\geq\half}=1$.

In the former case, the Bayes classifier $\thr{Q}$ is the constant predictor $(x,a)\mapsto 0$, thus $\acc{\thr{Q}}\leq\tau$ necessarily.
In the latter case, the alternative Bayes classifier $\thr{Q}^*$ (defined in Lemma~\ref{lem:alt-bayes}) is the constant predictor $(x,a)\mapsto1$, thus $\acc{\thr{Q}^*}\leq\tau$.
According to Lemma~\ref{lem:alt-bayes}, $\acc{\thr{Q}}=\acc{\thr{Q}^*}$, thus we may conclude $\acc{\thr{Q}}\leq\tau$ as well.

It follows that $\acc{\Q}\leq\acc{\thr{Q}}\leq\tau$ for all $\Q\in\setQ$ because $\thr{Q}$ has maximal accuracy in $\setQ$.

$(\Rightarrow)$ Suppose $\tau^{*}<1$.
We will suppose that the Bayes classifier $\thr{Q}$ is not trivially accurate and find a contradiction.

Suppose $\acc{\thr{Q}}-\tau=0$.
According to Lemmas~\ref{lem:acc-bayes} and \ref{lem:acc-constant} we may rewrite this as $\E{\left|Q-\half\right|-\left|\E Y-\half\right|}=0$.
Using the reverse triangle inequality, we conclude $\E{\left|Q-\E Y\right|}=0$, thus $Q=\E{Y}$ is constant.

If $\E Y\leq\half$, then $\P{Q\leq\half}=1$.
If $\E Y\geq\half$, then $\P{Q\geq\half}=1$.
In any case, we have $\tau^{*}=1$
which contradicts the initial supposition.

\end{proof}

%% file: body/lemmas.tex
\subsection{Some helper lemmas}
\begin{lemma}
\label{lem:err-exp}For every $\Q\in\setQ$, 
\[
\err{\Q}=\E{|\Q-Y|}
\]
\end{lemma}

\begin{proof}

Notice that $\PP{\Y\neq Y}{Y=1}=\EE{1-\Q}{Y=1}$ and $\PP{\Y\neq Y}{Y=0}=\EE{\Q}{Y=0}$.
In both cases, we may write $\PP{\Y\neq Y}{Y=y}=\EE{|Y-\Q|}{Y=y}$.

Hence, marginalizing over $Y$ we conclude $\P{\Y\neq Y}=\E{|Y-\Q|}$.

\end{proof}

\begin{lemma}\label{lem:vector-metrics}%
  (Vectorial metrics)
Using the notation of Definition~\ref{def:vector-setup}, we have
\[
  \begin{alignedat}{1}
    \err{\Q} &= \err{\VF} \\
    \oppDiff{\Q} &= \oppDiff{\VF}\\
  \end{alignedat}
  \]
\end{lemma}

\begin{proof}
For the error, we marginalize over $(X,A)$. Notice
\[\begin{alignedat}{1}
  &\PP{Y \neq \Y}{X=x_i,A=a_i} \\
  =& + (1-q(x_i,a_i)) \q(x_i,a_i)
  \\
  &+ q(x_i,a_i) (1-\q(x_i,a_i)) \\
  =& (1-\VQ_i) \frac{\VF_i}{\VP_i} + \VQ_i \frac{\VP_i-\VF_i}{\VP_i} \\
  =& \frac{\VQ_i \VP_i + \VF_i (1-2\VQ_i)}{\VP_i}\\
\end{alignedat}\]
Thus $\P{Y \neq \Y,X=x_i,A=a_i} = \VF_i \VP_i + \VF_i (1-2\VQ_i)$, and 
\[\begin{alignedat}{1}
  \err{f} &= \P{\Y \neq Y}\\
  &= \sum_{i=1}^n \P{\Y \neq Y,X=x_i,A=a_i}\\
  &= \inner{\VP}{\VQ} + \inner{\VF}{1-2Q}\\
  &= \err{\VF} \\
\end{alignedat}\]
For opportunity-difference, we also marginalize over $(X,A)$. Notice that
\[\begin{alignedat}{1}
  &\P{\Y=1, Y=1, X=x_i,A=a_i} \\
  =&\VP_i\, \EE{\Q Q}{X=x_i,A=a_i} \\
  =& \VP_i \frac{\VF_i}{\VP_i} \VQ_i = \VF_i \VQ_i, \\
\end{alignedat}\]
hence $\P{\Y=1,Y=1,X=x_i,A=a} = \VF_i \VQ^{(a)}_i$. In addition, $\P{Y=1,X=x_i,A=a} = \VP_i \VQ^{(a)}_i$ and
\[\begin{alignedat}{1}
  & \PP{\Y=1}{Y=1,A=a} \\
  =& \frac{\sum_{i=1}^n \P{\Y=1,Y=1,X=x_i,A=a}}{\sum_{i=1}^n \P{Y=1,X=x_i,A=a}} \\
  =& \frac{\inner{\VF}{\VQ^{(a)}}}{\inner{\VP}{\VQ^{(a)}}} \\
\end{alignedat}\]
Therefore
\[\begin{alignedat}{1}
  \oppDiff{\Q} =
  &+ \PP{\Y=1}{Y=1,A=1}\\
  &- \PP{\Y=1}{Y=1,A=0}\\
  =& \frac{\inner{\VF}{\VQ^{(1)}}}{\inner{\VP}{\VQ^{(1)}}}
  - \frac{\inner{\VF}{\VQ^{(0)}}}{\inner{\VP}{\VQ^{(0)}}}\\
  =& \oppDiff{\VF} \\
\end{alignedat}\]
\end{proof}

\begin{lemma}\label{lem:vector-symmetry}%
(Metrics symmetry)
Using the notation of Definition~\ref{def:vector-setup}, we have\[\begin{alignedat}{1}
  \err{\VP-\VF} &= 1 - \err{\VF} \\
  \oppDiff{\VP-\VF} &= - \oppDiff{\VF} \\
\end{alignedat}\]
\end{lemma}

\begin{proof}
According to Lemma~\ref{lem:vector-metrics}, opportunity-difference is a linear transformation. Since linear transformations preserve scalar multiplication and vector addition, it follows that $\oppDiff{\VP-\VF} = \oppDiff{\VP} - \oppDiff{\VF}$.
Moreover, since $\oppDiff{\VP}=1-1=0$, then $\oppDiff{\VP-\VF} = - \oppDiff{\VF}$.

According to the same lemma,  the error is an affine transformation with offset $\inner{\VP}{\VQ}$. Hence\[\begin{alignedat}{1}
  \err{\VP-\VF} &= \err{\VP} - \err{\VF} + \inner{\VP}{\VQ}\\
  &= 2\inner{\VP}{\VQ} - \inner{\VP}{1-2\VQ} - \err{\VF}\\
  &= \inner{\VP}{1} - \err{\VF}\\
  &= 1 - \err{\VF}\\
\end{alignedat}\]
because $\sum_{i=1}^n \VP_i = 1$.
\end{proof}

\begin{lemma}
\label{lem:acc-bayes}(Bayes accuracy)
\[
\acc{\thr{Q}}=\half+\E{|Q-\half|}
\]
\end{lemma}

\begin{proof}

Out of Lemma~\ref{lem:err-exp}, we know $\err{\thr{Q}}=1-\E \epsilon$
where $\epsilon\eqdef|\thr{Q}-Y|$.
Let us condition on $Q<\half$ and $Q\geq\half$ separately
(whenever these events have possible probabilities).

For $Q<\half$, we have $\EE \epsilon{Q<\half}=\EE Y{Q<\half}=\EE{Q}{Q<\half}$
and $Q=\half-(\half-Q)$. For $Q\geq\half$, we
have $\EE \epsilon{Q\geq\half}=\EE{1-Y}{Q\geq\half}=\EE{1-Q}{Q\geq\half}$
and $1-Q=\half-(Q-\half)$.

These cases partition $\Pomega$ and in both cases we have $\E \epsilon=\half-\E{\half-Q}$.
It follows that $\err{\thr{Q}}=\half-\E{|Q-\half|}$.

\end{proof}
\begin{lemma}
\label{lem:err-half}(Uniform case) Let $\Q\in\setQ$ and $\epsilon\eqdef|\Q-Y|$
be the random variable of the error of $\Q$ (according to Lemma~\ref{lem:err-exp}).
If $\P{Q=\half}>0$, then 
\[
\EE{\epsilon}{Q=\half}=\half
\]
\end{lemma}

\begin{proof}

Define $r\eqdef\E{\Q}$.
Let us condition on $Y=0$ and $Y=1$ separately. For $Y=0$, we have $\EE{\epsilon}{Q=\half,Y=0}=r$,
and for $Y=1$, we have $\EE{\epsilon}{Q=\half,Y=1}=1-r$.

Since $\PP{Y=y}{Q=\half}=\half$, we can compute the marginal
as
\[
\EE{\epsilon}{Q=\half}=(\half)\compact{(r+1-r)}=\half
\]

\end{proof}
\begin{lemma}\label{lem:alt-bayes}
(Alternative Bayes)
The \emph{alternative Bayes classifier} $\thr{Q}$ given by $\one{q(x,a)\geq\half}$ ($\geq$ instead of $>$) has also maximal accuracy.
\end{lemma}

\begin{proof}

We will prove that $\err{\thr{Q}}=\err{\thr{Q}^*}$. Following Lemma~\ref{lem:err-exp},
let $\epsilon\eqdef|\thr{Q}-Y|$ and $\epsilon^{*}\eqdef|\thr{Q}^*-Y|$.

Conditioned to $Q\neq\half$ we have $\thr{Q}=\thr{Q}^*$ from their
definitions, and thus also $\EE{\epsilon-\epsilon^{*}}{Q\neq\half}=0$.
It suffices to check the complement event $Q = \half$.
Suppose $\P{Q=\half}>0$.
Conditioned to $Q=\half$, Lemma~\ref{lem:err-half} implies that
$\EE{\epsilon-\epsilon^{*}}{Q=\half}=\half-\half=0$.

Hence $\E{\epsilon}=\E{\epsilon^{*}}$, i.e. $\err{\thr{q}}=\err{\thr{q}^*}$.

\end{proof}
\begin{lemma}\label{lem:acc-constant}
(Trivial error as an expectation)
\[
\tau=\half+|\E Y-\half|
\]
\end{lemma}

\begin{proof}

The constant $0$ predictor ($\hat 0$) has error $\E Y$, while the constant
$1$ predictor ($\hat 1$) has error $1-\E Y$.
We can rewrite these quantities
respectively as $\half-\left(\half-\E Y\right)$ and $\half+\left(\half-\E Y\right)$,
whose maximum is $\tau=\half+|\half-\E Y|$.

\end{proof}

\begin{lemma}\label{lem:plane1}
Let $\VP,\VQ\in(0,1)^3$, with $\VQ_1<\half$ and $\VQ_2 - \VQ_1 > 0$ (as in Theorem~\ref{thm:plane}).

If $\VP$ and $\VQ$ satisfy also $\VQ_3+\VQ_1 \geq 1$ and $\VP_1 \VQ_1 + \VP_2 \VQ_2 < \VP_3 \VQ_1$, then $\inner{\VP}{2\VQ-1}>\inner{\VZ}{2\VQ-1}$.
\end{lemma}

\begin{proof}
We have the following equivalences and at the end an implication.
\[\begin{alignedat}{1}
  &\inner{2\VQ-1}{\VP}-\inner{2\VQ-1}{\VZ} > 0\\
  \equiv& \inner{2\VQ-1}{\VP-\VZ}>0\\
  \equiv& \compact{(2\VQ_1-1)\VP_1 + (2\VQ_3-1) \VP_3\frac{\VP_1 \VQ_1}{\VP_1 \VQ_1+\VP_2 \VQ_2}>0}\\
  \equiv& \compact{(2\VQ_1-1)(\VP_1 \VQ_1+\VP_2 \VQ_2) + (2\VQ_3-1) \VP_3 \VQ_1>0}\\
  \equiv& \compact{(1-2\VQ_1)(\VP_1 \VQ_1+\VP_2 \VQ_2) < (2\VQ_3-1) \VP_3 \VQ_1}\\
  \Leftarrow & \compact{(1-2\VQ_1 \leq 2\VQ_3-1) \wedge (\VP_1 \VQ_1+\VP_2 \VQ_2 < \VP_3 \VQ_1)}\\
\end{alignedat}\]

It is given that $\VQ_3+\VQ_1 \geq 1$ and $\VP_1 \VQ_1 + \VP_2 \VQ_2 < \VP_3 \VQ_1$, which are equivalent to the last two inequalities.
Thus, they imply that $\inner{2\VQ-1}{\VP}-\inner{2\VQ-1}{\VZ} > 0$.

\end{proof}

\begin{lemma}
\label{lem:plane2}(Complementary part of Theorem~\ref{thm:plane})
Algorithm~\ref{alg:plane} is correct.
\end{lemma}

\begin{proof}
\setcounter{proofpart}{0}
We will prove $a<b$, $\VP_2\in(0,1)$ and the fulfillment of constraints \ref{enu:constr-constant} and \ref{enu:constr-P}.

\proofpart{Proof that $a<b$.}

Recall $a = \max\{(1-\VP_3)\VQ_1, \half - \VP_3 \VQ_3\}$ and $b = \min\{(1-\VP_3)\VQ_2, \VP_3 \VQ_1\}$.

\begin{enumerate}
  \item Since $\VQ_1<\half<\VQ_2$ and $\VP_3\in(0,1)$, then $(1-\VP_3)\VQ_1 < (1-\VP_3) \VQ_2$.
  \item Since $\VP_3\in(\half, 1)$, then $(1-\VP_3)\VQ_1 < \VP_3 \VQ_1$.
  \item Since $\VP_3\in(0,1)$ and $\VQ_3\in(\half,1)$, then $\VP_3 (\VQ_2-\VQ_3) < 1 \cdot (\VQ_2 - \half)$, or equivalently, $\half - \VP_3 \VQ_3 < (1-\VP_3) \VQ_2$.
  \item Since $\VP_3 > \frac{1}{2(\VQ_1+\VQ_3)}$ then $\half-\VP_3 \VQ_3 < \VP_3 \VQ_1$.
\end{enumerate}

Since the inequalities hold for all available choices for $a$ and $b$, then, in general, $a < b$ holds.

\proofpart{Proof that $\VP_2\in(0,1)$.}\\*
We know $c > \VQ_1 (1-\VP_3)$ and $c<\VQ_2 (1-\VP_3)$.
These inequalities imply that $c - \VQ_1(1-\VP_3) \in (0, \VQ_2-\VQ_1)$, hence also that $\VP_2 \in (0,1)$.

\proofpart{Constraint~\ref{enu:constr-constant} is satisfied.}
\\*
Since $\VP_1 + \VP_2 = 1-\VP_3$ and $\VQ_2>\VQ_1$, then the term $\VP_1 \VQ_1 + \VP_2 \VQ_2$ is minimal when $\VP_1=1-\VP_3$ and $\VP_2=0$. Thus,
\[\begin{alignedat}{1}
  & \VP_1 \VQ_1 + \VP_2 \VQ_2 + \VP_3 \VQ_3\\
  &\geq  (1-\VP_3) \VQ_1 + \VP_3 \VQ_3\\
  &= \VQ_1 + \VP_3 (\VQ_3-\VQ_1)\\
  &> \VQ_1 + \frac{\VQ_3-\VQ_1}{2}\\
  &= \frac{\VQ_3+\VQ_1}{2} \geq \half.\\
\end{alignedat}\]

\proofpart{Constraint~\ref{enu:constr-P} is satisfied.}\\*
Since $b \leq \VP_3 \VQ_1$, then $\VP_2 (\VQ_2-\VQ_1) < \VP_3 \VQ_1 - \VQ_1(1 - \VP_3)$.
From this inequality, we may derive constraint~\ref{enu:constr-P} as follows.
\[\begin{alignedat}{1}
  \VP_2 (\VQ_2-\VQ_1) &< \VP_3 \VQ_1 - \VQ_1(1 - \VP_3)\\
  \VP_2 \VQ_2 &< (2 \VP_3-1+\VP_2) \VQ_1\\
  \VP_2 \VQ_2 &< \VP_3 \VQ_1 - \VP_1 \VQ_1\\
  \VP_1 \VQ_1 + \VP_2 \VQ_2 &< \VP_3 \VQ_1.\\
\end{alignedat}\]
  
\end{proof}

\begin{lemma}\label{lemma:EQQ}
Assume $\P{Y=1,A=a}>0$ for each $a\in\imA$. For any predictor $\Q$, we have
\[
    \PP{\hat Y=1}{Y=1,A=a} =
    \frac{\EE{\Q Q}{A=a}}{\EE{Q}{A=a}}
\]
hence also
\[
  \oppDiff{\Q} =
  \frac{\EE{\Q Q}{A=1}}{\EE{Q}{A=1}}
  -\frac{\EE{\Q Q}{A=0}}{\EE{Q}{A=0}}
\]
As an additional consequence, by considering the symmetric predictor $1-\Q$, it is also true that
\[
    \PP{\hat Y=0}{Y=1,A=a} =
    \frac{\EE{(1-\Q) Q}{A=a}}{\EE{Q}{A=a}}
\]
\end{lemma}
\begin{proof}
Indeed, applying repetitively the Bayes rule, we get
\[
\begin{aligned}
  &\PP{\Y=1}{Y=1,A=a}\\
  &=\frac{\P{\Y=1,Y=1,A=a}}{\P{Y=1,A=a}}\\
  &=\frac{\P{A=a}\EE{\Q Q}{A=a}}{\P{Y=1,A=a}}\\
  &=\frac{\EE{\Q Q}{A=a}}{\PP{Y=1}{A=a}}\\
  &=\frac{\EE{\Q Q}{A=a}}{\EE{Q}{A=a}}
\end{aligned}
\]
The second equality holds because $(Y,\Y)\perp A \mid (Q,\Q)$ and $Y\perp \Y \mid (Q,\Q)$.
\end{proof}